\documentclass[runningheads,a4paper]{llncs}

\usepackage[cmex10]{amsmath}
\usepackage{amssymb}
\usepackage{llncsdoc}
\usepackage{mathptmx}       
\usepackage{helvet}         
\usepackage{courier}        
\usepackage{makeidx}         
\usepackage[pdftex]{graphicx}
\usepackage{epstopdf}
\DeclareGraphicsExtensions{.pdf,.jpeg,.png,.eps}
\usepackage{multirow}        
\usepackage[bottom]{footmisc}

\usepackage{url}
\usepackage[linesnumbered,ruled,vlined]{algorithm2e}
\usepackage{array}
\usepackage{booktabs}
\usepackage[caption=false,font=footnotesize]{subfig}
\usepackage{xcolor,colortbl}
\usepackage{bm}

\newcommand{\norm}[2][2]{\left\lVert #2 \right\rVert_{#1}}
\newcommand{\vect}[1]{\mathbf{#1}}
\newcommand{\ra}[1]{\renewcommand{\arraystretch}{#1}}
\DeclareMathSymbol{\R}{\mathalpha}{AMSb}{"52}

\definecolor{Gray}{gray}{0.85}
\definecolor{LightCyan}{rgb}{0.88,1,1}

\begin{document}
\mainmatter 

\title{Learning activation functions from data \\ using cubic spline interpolation}

\author{Simone Scardapane \and Michele Scarpiniti \and Danilo Comminiello \and Aurelio Uncini}
\titlerunning{Learning activation functions from data}
\authorrunning{Scardapane et al.}
\institute{Department of Information Engineering, Electronics and Telecommunications (DIET), \\ ``Sapienza'' University of Rome, \\ Via Eudossiana 18, 00184, Rome. \\Email: \{simone.scardapane, michele.scarpiniti, danilo.comminiell\}@uniroma1.it; aurel@ieee.org}

\maketitle

\begin{abstract}
Neural networks require a careful design in order to perform properly on a given task. In particular, selecting a good activation function (possibly in a data-dependent fashion) is a crucial step, which remains an open problem in the research community. Despite a large amount of investigations, most current implementations simply select one \textit{fixed} function from a small set of candidates, which is not adapted during training, and is shared among all neurons throughout the different layers. However, neither two of these assumptions can be supposed optimal in practice. In this paper, we present a principled way to have data-dependent adaptation of the activation functions, which is performed \textit{independently} for each neuron. This is achieved by leveraging over past and present advances on cubic spline interpolation, allowing for local adaptation of the functions around their regions of use. The resulting algorithm is relatively cheap to implement, and overfitting is counterbalanced by the inclusion of a novel damping criterion, which penalizes unwanted oscillations from a predefined shape. Preliminary experimental results validate the proposal.
\end{abstract}

\section{Introduction}
\label{sec:intro}

Neural networks (NNs) are extremely powerful tools for approximating complex nonlinear functions \cite{haykin2009neural}. The nonlinear behavior is introduced in the NN architecture by the elementwise application of a given nonlinearity, called the activation function (AF), at every layer. Since AFs are crucial to the dynamics and computational power of NNs, the history of the two over the last decades is deeply connected \cite{schmidhuber2015deep}. As an example, the use of \textit{differentiable} AFs was one of the major breakthroughs in NNs, leading directly to the back-propagation algorithm. More recently, progress on piecewise linear functions was shown to facilitate backward flow of information for training very deep networks \cite{glorot2010understanding}. At the same time, it is somewhat surprising that the vast majority of NNs only use a small handful of fixed functions, to be hand-chosen by the practitioner before the learning process. Worse, there is no principled reason to believe that a `good' nonlinearity might be the same across all layers of the network, or even across neurons in the same layer.

This is shown clearly in a recent work by Agostinelli \textit{et al.} \cite{agostinelli2014learning}, where every neuron in a deep network was endowed with an adaptable piecewise linear function with possibly different parameters, concluding that ``\textit{the standard one-activation-function-fits-all approach may be suboptimal}'' in current practice. Experiments in AF adaptation have a long history, but they have never met a wide applicability in the field. The simplest approach is to parameterize each sigmoid function in the network by one or more `shape' parameters to be optimized, such as in the seminal 1996 paper by Chen and Chang \cite{chen1996feedforward} or the later work by Trentin \cite{trentin2001networks}. Along a similar line, one may consider the use of polynomial AFs, wherein each coefficient of the polynomial is adapted by gradient descent \cite{piazza1992artificial}. Additional investigations can be found in \cite{zhang2002neuron,goh2003recurrent,chandra2004activation,ma2005constructive,lin2013network}. One strong drawback of these approaches is that the parameters involved affect the AF \textit{globally}, such that a change in one region of the function may be counterproductive on a different, possibly faraway, region.

Several years ago, an alternative approach was introduced by using spline interpolating functions as AFs \cite{vecci1998learning,guarnieri1999multilayer}, resulting in what was called a spline AF (SAF). Splines are an attractive choice for interpolating unknown functions, since they can be described by a small amount of parameters, yet each parameter has a local effect, and only a fixed number of them is involved every time an output value is computed \cite{wahba1990spline}. The original works in \cite{vecci1998learning,guarnieri1999multilayer} had two main drawbacks that prevented a wider use of the underlying theory. First, SAFs were only investigated in an online setting, where updates are computed one sample at a time. Whether an efficient implementation is possible (and feasible) also for batch (or mini-batch) settings was not shown. Secondly, the obtained SAFs had a tendency to overfit training data, resulting in oscillatory behaviors which hindered performance. Inspired by recent successes in the field of nonlinear adaptive filtering \cite{scarpiniti2013nonlinear,scarpiniti2016steadystate}, our aim in this paper is two-fold. On one hand, we provide a modern introduction to the use of SAFs in neural networks, with a particular emphasis on their efficient implementation in the case of batch (or mini-batch) training. Our treatment clearly shows that the major problem in their implementation, which is evident from the discussion above, is the design of an efficient way to regularize their control points. In this sense, as a second contribution we provide a simple (yet effective) `damping' criterion to prevent unwanted oscillations in the testing phase, which penalizes deviations from the original points in terms of $\ell_2$ norm. A restricted set of experiments shows that the resulting formulation is able to achieve a lower test error than a standard NN with fixed AFs, while at the same time learning non-trivial activations with different shapes across different neurons.

The rest of the paper is organized as follows. Section \ref{sec:spline_activation_function} presents the basic theory of SAFs for the case of a single neuron. Section \ref{sec:neural_networks_with_saf_neurons} extends the treatment to the case of a NN with one hidden layer, by deriving the gradient equations for the SAFs parameters in the internal layer. Then, Section \ref{sec:experimental_results} goes over the experimental results, while we conclude with some final remarks in Section \ref{sec:conclusion}.

\section{The spline activation function}
\label{sec:spline_activation_function}

We begin our treatment of SAFs with the simplest case of a single neuron endowed with a flexible AF (see \cite{vecci1998learning,scarpiniti2013nonlinear} for additional details). Given a generic input $\vect{x} \in \R^D$, the output of the SAF is computed as:
\begin{align}
s & = \vect{w}^T\vect{x} \,, \\
y & = \varphi(s; \vect{q}) \,, \label{eq:saf_output_generic}
\end{align}
where $\vect{w} \in \R^D$ (we suppose that an eventual bias term is added directly to the input vector), and the AF $\varphi(\cdot)$ is parameterized by a vector $\vect{q} \in \R^Q$ of internal parameters, called \textit{knots}. The knots are a sampling of the AF values over $Q$ representative points spanning the overall function. In particular, we suppose the knots to be uniformly spaced, i.e. $q_{i+1} = q_{i} + \Delta x$, for a fixed $\Delta x \in \R$, and symmetrically spaced around the origin. Given $s$, the output is computed by spline interpolation over the closest knot and its $P$ rightmost neighbors. The common choice $P=3$, which we adopt in this paper, corresponds to \textit{cubic} interpolation, and it is generally a good trade-off between locality of the output and interpolating precision.

Given the index $i$ of the closest knot, we can define the normalized abscissa value between $q_{i}$ and $q_{i+1}$ as:
\begin{equation}
u = \frac{s}{\Delta x} - \left\lfloor \frac{s}{\Delta x} \right\rfloor \,.
\label{eq:u}
\end{equation}
where $\lfloor \cdot \rfloor$ is the floor operator. From $u$ we can compute the normalized reference vector $\vect{u} = \left[ u^P \; u^{P-1} \ldots u \; 1 \right]^T$, while from $i$ we can extract the relevant control points $\vect{q}_{i} = \left[ q_{i} \; q_{i+1} \ldots q_{i+P} \right]^T$. We refer to the vector $\vect{q}_{i}$ as the $i$th \textit{span}. The output \eqref{eq:saf_output_generic} is then computed as:
\begin{equation}
y = \varphi(s) = \vect{u}^T\vect{B}\vect{q}_{i} \,,
\label{eq:saf_output}
\end{equation}
where $\vect{B} \in \R^{\left(P+1\right) \times \left(P+1\right)}$ is called the spline basis matrix. In this work, we use the Catmull-Rom (CR) spline with $P=3$, given by:
\begin{equation}
\vect{B} = \frac{1}{2} 
\begin{bmatrix}
	-1 & 3 & -3 & 1 \\
	2 & -5 & 4 & -1 \\
	-1 & 0 & 1 & 0 \\
	0 & 2 & 0 & 0
\end{bmatrix} \,.
\label{eq:catmulrom_basis_matrix}
\end{equation}
Different bases give rise to alternative interpolation schemes, e.g. a spline defined by a CR basis passes through all the control points, but its second derivative is not continuous.

Apart from the locality of the output, SAFs have two additional interesting properties. First, the output in \eqref{eq:saf_output} is extremely efficient to compute, involving only vector-matrix products of very small dimensionality. Secondly, derivatives with respect to the internal parameters are equivalently simple and can be written down in closed form. In particular, the derivative of the nonlinearity $\varphi(s)$ with respect to the input $s$ is given by:
\begin{equation}
\frac{\partial \varphi(s)}{\partial s} = \varphi'(s) = \frac{\partial \varphi(s)}{\partial u}\cdot\frac{\partial u}{\partial s} = \left(\frac{1}{\Delta x}\right) \dot{\vect{u}} \vect{B}\vect{q}_{i} \,,
\label{eq:varphi_derivative}
\end{equation}
where:
\begin{equation}
\dot{\vect{u}} = \frac{\partial \vect{u}}{\partial u} = \left[ Pu^{P-1} \; (P-1)u^{P-2} \ldots 1 \; 0 \right]^T \,.
\label{eq:u_dot}
\end{equation}
Given this, the derivative of the SAF output $y$ with respect to $\vect{w}$ is straightforward:
\begin{align}
\frac{\partial \varphi(s)}{\partial \vect{w}} & = \varphi'(s)\cdot\frac{\partial s}{\partial \vect{w}} = \varphi'(s)\vect{x} \,,
\label{eq:saf_derivative_linear}
\end{align}
Similarly, for $\vect{q}_i$ we obtain:
\begin{align}
\frac{\partial \varphi(s)}{\partial \vect{q}_{i}} & = \vect{B}^T \vect{u} \,.
\label{eq:saf_derivative_nonlinear}
\end{align}
while we have $\frac{\partial \varphi(s)}{\partial q_k} = 0$ for any element $q_k$ outside the current span $\vect{q}_i$.

\section{Designing networks with SAF neurons}
\label{sec:neural_networks_with_saf_neurons}

\subsection{Computing outputs and inner derivatives}
\label{subsec:computing_outputs_and_inner_derivatives}

Now we consider the more elaborate case of a single hidden layer NN, with a $D$-dimensional input, $H$ neurons in the hidden layer, and $O$ output neurons.\footnote{We note that the following treatment can be extended easily to the case of a network with more than one hidden layer. However, restricting it to a single layer allow us to keep the discussion focused on the problems/advantages arising in the use of SAFs. We leave this extension to a future work.} Every neuron in the network uses a SAF with possibly different adaptive control points, which are set independently during the training process. For easiness of computation, we suppose that the sampling set of the splines is the same for every neuron (i.e., each neuron has $Q$ points equispaced according to the same $\Delta x$), and we also have a single shared basis matrix $\vect{B}$. The forward phase of the network is similar to that of a standard network. In particular, given the input $\vect{x}$, we first compute the output of the $i$th hidden neuron, $i=1,\ldots,H$, as:
\begin{equation}
h_i = \varphi(\vect{w}_{h_i}^T\vect{x}; \vect{q}_{h_i}) \,.
\end{equation}
These are concatenated in a single vector $\vect{h}=\left[ h_1, \ldots, h_H, 1 \right]^T$, and the $i$th output of the network, $i=1,\ldots,O$, is given by:
\begin{equation}
f_i(\vect{x}) = y_i = \varphi(\vect{w}_{y_i}^T\vect{h}; \vect{q}_{y_i}) \,.
\end{equation}
The derivatives with respect to the parameters $\left\{\vect{w}_{y_i}, \vect{q}_{y_i}\right\}$, $i=1,\ldots,O$ can be computed directly with \eqref{eq:saf_derivative_linear}-\eqref{eq:saf_derivative_nonlinear}, substituting $\vect{x}$ with $\vect{h}$. By back-propagation, the derivative of the $i$th output with respect to the $j$th (inner) weight vector $\vect{w}_{h_j}$ is similar to a standard NN:
\begin{equation}
\frac{\partial y_j}{\partial \vect{w}_{h_i}} = \varphi'(s_{y_j})\cdot\varphi'(s_{h_i})\cdot\lfloor\vect{w}_{h_i}\rfloor_j\cdot\vect{x} \,,
\end{equation}
where with a slight abuse of notation we let $s_{y_j}$ denote the activation of the $j$th output (and similarly for $s_{h_i}$), $\lfloor\cdot\rfloor_j$ extracts the $j$th element of its input vector, and the two $\varphi'(\cdot)$ are given by \eqref{eq:varphi_derivative}. For the derivative of the control points of the $i$th hidden neuron, denote by $\vect{q}_{h_i,k}$ the currently active span, and by $\vect{u}_{h_i}$ the corresponding reference vector. The derivative with respect to the $j$th output is then given by:
\begin{equation}
\frac{\partial y_j}{\partial \vect{q}_{h_i,k}} = \varphi'(s_{y_j})\cdot\lfloor\vect{w}_{h_i}\rfloor_j\cdot\vect{B}^T\vect{u}_{h_i}\,.
\label{eq:saf_derivative_nonlinear_hidden}
\end{equation}

\subsection{Initialization of the control points}
An important aspect that we have not discussed yet is how to properly initialize the control points. One immediate choice is to sample their values from an AF which is known to work well on the given problem, e.g. a hyperbolic tangent. In this way, the network is guaranteed to work similarly to a standard NN in the initial phase of learning. Additionally, we have found good improvements in error by adding Gaussian noise $\mathcal{N}(0, \sigma^2)$ with small variance $\sigma^2$ to a randomly chosen subset of control points (around $5\%$ in our experiments). This provides a good variability in the beginning, similarly to how connections are set close to (but not identically equal to) zero during initialization.

\subsection{Choosing a training criterion}

Suppose we are provided with a training set of $N$ input/output pairs in the form $\left\{ \vect{x}_i, \vect{d}_i \right\}_{i=1}^N$. For simplicity of notation, we denote by $\vect{w}$ the concatenation of all weight vectors $\left\{\vect{w}_{h_i}\right\}$ and $\left\{\vect{w}_{y_i}\right\}$, and by $\vect{q}$ a similar concatenation of all control points. Training can be formulated as the minimization of the following cost function:
\begin{align}
J(\vect{w}, \vect{q}) = \frac{1}{N}\sum_{i=1}^N L(\vect{d}_i, \vect{f}(\vect{x}_i)) + \lambda_w R_w(\vect{w}) + \lambda_q R_q(\vect{q}) \,,
\end{align}
where $L(\cdot, \cdot)$ is an error function, while $R_w(\cdot)$ and $R_q(\cdot)$ provide meaningful regularization on the two set of parameters. The first two terms are well-known in the neural network literature \cite{haykin2009neural}, and they can be set accordingly. Particularly, in our experiments we consider a squared error term $L(\vect{d}_i, \vect{f}(\vect{x}_i)) = \norm{\vect{d}_i - \vect{f}(\vect{x}_i)}^2$, and $\ell_2$ regularization on the weights $R_w(\vect{w}) = \norm{\vect{w}}^2$. The derivatives of $L(\cdot, \cdot)$ can be computed straightforwardly with the formulas presented in Section \ref{subsec:computing_outputs_and_inner_derivatives}. 

The term $R_q(\vect{q})$ is used to avoid overfitted solutions for the control points. In fact, its presence is the major difference with respect to previous attempts at implementing SAFs in neural networks \cite{vecci1998learning}, wherein overfitting was counterbalanced by choosing a large value for $\Delta x$, which in a way goes outside the philosophy of spline interpolation itself. At the same time, choosing a proper form for the regularization term is non-trivial, as the term should be cheap to compute, and it should introduce just as much \textit{a priori} information as needed, without hindering the training process. Most of the literature on regularizing $\vect{w}$ cannot be used here, as the corresponding formulations do not make sense in the context of spline interpolation. As an example, simply penalizing the $\ell_2$ norm of $\vect{q}$ leads to functions close to the zero function, while imposing sparsity is also meaningless.

For the purpose of this paper, we consider the following `damping' criterion:
\begin{equation}
R_q(\vect{q}) = \norm{\vect{q} - \vect{q}_o}^2 \,,
\end{equation}
where $\vect{q}_o$ represents the initial values for the control points, as discussed in the previous section (without considering additional noise). The criterion makes intuitive sense as follows: while for $\vect{w}$ we wish to penalize unwanted deviations from very small weights (which can be justified with arguments from learning theory), in the case of $\vect{q}$ we are interested in penalizing changes with respect to a `good' function parameterized by the initial control points $\vect{q}_o$, namely one of the standard AFs used in NN training. In fact, setting a value for $\lambda_q$ very high essentially deactivates the adaptation of the control points. Clearly, other choices are possible, and in this sense this paper serves as a starting point for further investigations towards this objective. As an example, we may wish to penalize first (or second) order derivatives of the splines in order to force a desired level of smoothness \cite{wahba1990spline}.

\subsection{Remarks on the implementation}
In order to be usable in practice, SAFs require an efficient implementation to compute outputs and derivatives concurrently for the entire training dataset or, alternatively, for a properly chosen mini-batch (in the case of stochastic optimization algorithms). To begin with, we underline that the equations for the reference vector (see \eqref{eq:u}) do not depend on the specific neuron, and for this reason they can easily be vectorized layer-wise on most numerical algebra libraries to obtain all vectors concurrently. Additionally, the indexes and relative terms $\vect{B}\vect{q}_i$ in \eqref{eq:saf_output} can be cached during the forward pass, to be reused during the computation of the derivatives. In this sense, the outputs of a layer \textit{and} its derivatives can be computed by one $4\times4$ matrix-vector computation, and three $4$-dimensional inner products, which have to be repeated for every pair input/neuron. In our experience, the cost of a relatively well-optimized implementation does not exceed twice that of a standard network for medium-sized batches, where the most onerous operation is the reshaping of the gradients in \eqref{eq:saf_derivative_nonlinear} and \eqref{eq:saf_derivative_nonlinear_hidden} into a single vector of gradients relative to the global vector $\vect{q}$.

\section{Experimental results}
\label{sec:experimental_results}

\subsection{Experimental setup}

To evaluate the preliminary proposal, we consider two simple regression benchmarks for neural networks, the `chemical' dataset (included among MATLAB's testbeds for function fitting), and the `California Housing'.\footnote{\url{http://www.dcc.fc.up.pt/~ltorgo/Regression/cal_housing.html}} They have respectively $498$ and $20640$ examples, and $8$ numerical features. Inputs are normalized in the $\left[-1,+1\right]$ range, while outputs are normalized in the $\left[-0.5,+0.5\right]$ range. We compare a NN with $5$ hidden neurons and $\tanh(\cdot)$ AFs (denoted as `Standard' in the results), and a NN with the same number of neurons and SAF nonlinearities. The weight vector $\vect{w}$ is initialized with the method described in \cite{glorot2010understanding}. Each SAF is initialized from a $\tanh(\cdot)$ nonlinearity, and control points are defined in the $\left[-2,+2\right]$ range with $\Delta x=0.2$, which is a good compromise between locality of the SAFs and the overall number of adaptable parameters. For the first scenario, $\lambda_q$ is kept to a small value of $10^{-5}$. For each experiment, a random $30\%$ of the dataset is kept for testing, and results are averaged over $15$ different splits to average out statistical effects. Error is computed with the Normalized Root Mean-Squared Error (NRMSE). The optimization problems are solved using a freely available MATLAB implementation of the Polack-Ribiere variant of the nonlinear conjugate gradient optimization algorithm by C.E. Rasmussen. \cite{rasmussen2006gaussian}.\footnote{\url{http://learning.eng.cam.ac.uk/carl/code/minimize/}} The optimization process is allowed $1500$ maximum iterations. MATLAB code for the experiments is also available on the web.\footnote{[The URL has been hidden for the review process.]} We briefly remark that the MATLAB library, apart from repeating the experiments presented here, is also designed to handle networks with more than a single hidden layer, and implements the ADAM algorithm \cite{kingma2014adam} for stochastic training in case of a larger dataset.

\subsection{Scenario $1$: strong underfitting}

\begin{table}[t]
\ra{1.2}
\small
\centering
\caption{Average results for scenario $1$ ($\lambda_w=1$), together with one standard deviation.}\vspace{1em}
\begin{tabular}{llll}
\toprule
Dataset & Nonlinearity & Tr. RMSE & T.st NRMSE \\
\midrule
\multirow{2}{*}{Chemical} & Standard & $1.00 \pm 0.00$ & $1.00 \pm 0.01$ \\
						  & SAF      & $\mathbf{0.29 \pm 0.02}$ & $\mathbf{0.31 \pm 0.02}$ \\
\midrule
\multirow{2}{*}{Calhousing} & Standard & $1.02 \pm 0.00$ & $1.01 \pm 0.01$ \\
							& SAF	   & $\mathbf{0.56 \pm 0.01}$ & $\mathbf{0.57 \pm 0.02}$ \\
\bottomrule
\end{tabular}
\vspace{0.5em}
\label{tab:scenario_1_results}
\end{table}

As a first example, we consider a scenario of strong underfitting, wherein the user has misleadingly selected a very large value of $\lambda_w=1$, leading in turn to extremely small values for the elements of $\vect{w}$ after training. Results in terms of training and test RMSE are provided in Tab. \ref{tab:scenario_1_results}. Since the activations of the NN tend to be very close to $0$ (where the hyperbolic tangent operates in an almost-linear regime), standard NNs have a constant zero output, leading to a RMSE of $1$. Nonetheless, SAF networks are able to reach a very satisfactory level of performance, which in the first case is almost comparable to that of a fully optimized network (see the following section). 

\begin{figure}
	\centering
	\subfloat{\includegraphics{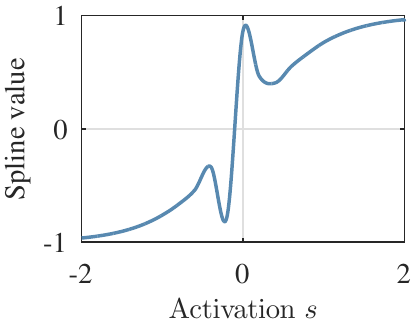}%
		\label{fig:scenario_1_chemical_spline_1}}
	\subfloat{\includegraphics{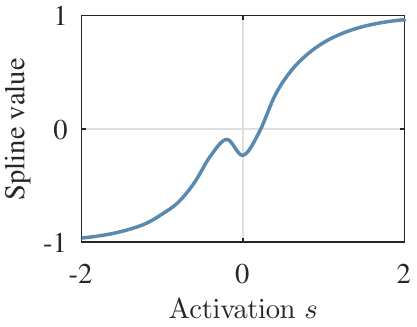}%
		\label{fig:scenario_1_chemical_spline_2}}
	\vfill
	\subfloat{\includegraphics{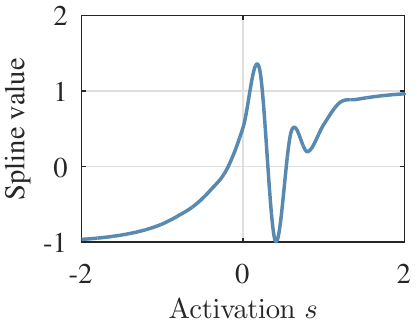}%
		\label{fig:scenario_1_chemical_spline_3}}
	\subfloat{\includegraphics{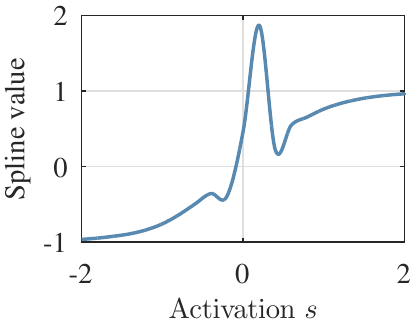}%
		\label{fig:scenario_1_chemical_spline_4}}
	\caption{Non-trivial representative SAFs after training for scenario $1$.}
	\label{fig:scenario_1_chemical}
\end{figure}

To show the reasons for this, we have plotted four representative nonlinearities after training in Fig. \ref{fig:scenario_1_chemical}. It is easy to see that the nonlinearities have adapted to act as `amplifiers' for the activations in their operating regime, with mild and strong peaks around $0$. Of particular interest is the fact that the resulting SAFs need not be perfectly centered around 0 (e.g. Fig. \ref{fig:scenario_1_chemical_spline_3}), or even symmetrical around the $y$-axis (e.g. Fig. \ref{fig:scenario_1_chemical_spline_4}). In fact, the splines are able to efficiently counterbalance a bad setting for the weights, with behaviors which would be very hard (or close to impossible) using standard setups with fixed, shared, mild nonlinearities.

\subsection{Scenario $2$: well-optimized parameters}

In our second scenario, we consider a similar comparison with respect to before, but we fine-tune the parameters of the two methods using a grid-search with a $3$-fold cross-validation on the training data as performance measure. Both $\lambda_w$ and $\lambda_q$ (only for the proposed algorithm) are searched in an exponential interval $2^{j}$, with $j=-10, \ldots, 5$. Optimal parameters found by the grid-search are listed in Table \ref{tab:scenario_2_optimal_parameters}, while results in terms of training and test NRMSE are collected in Table \ref{tab:scenario_2_results}.

\begin{table}[t]
\ra{1.2}
\small
\centering
\caption{Optimal parameters (averaged over the runs) found by the grid-search procedure for scenario $2$.}\vspace{1em}
\begin{tabular}{llll}
\toprule
Dataset & Nonlinearity & $\lambda_w$ & $\lambda_q$ \\
\midrule
\multirow{2}{*}{Chemical} & Standard & $10^{-3}$ & --- \\
						  & SAF      & $10^{-2}$ & $10^{-4}$ \\
\midrule
\multirow{2}{*}{Calhousing} & Standard & $10^{-4}$ & --- \\
						  & SAF      & $10^{-3}$ & $10^{-4}$ \\
\bottomrule
\end{tabular}
\vspace{0.5em}
\label{tab:scenario_2_optimal_parameters}
\end{table}

\begin{table}[t]
\ra{1.2}
\small
\centering
\caption{Average results for scenario $2$ (fine-tuning for parameters), together with one standard deviation.}\vspace{1em}
\begin{tabular}{llll}
\toprule
Dataset & Nonlinearity & Tr. RMSE & T.st NRMSE \\
\midrule
\multirow{2}{*}{Chemical} & Standard & $0.32 \pm 0.01$ & $0.32 \pm 0.02$ \\
						  & SAF      & $\mathbf{0.26 \pm 0.01}$ & $\mathbf{0.28 \pm 0.02}$ \\
\midrule
\multirow{2}{*}{Calhousing} & Standard & $0.55 \pm 0.01$ & $0.55 \pm 0.01$ \\
							& SAF	   & $\mathbf{0.51 \pm 0.02}$ & $\mathbf{0.51 \pm 0.02}$ \\
\bottomrule
\end{tabular}
\vspace{0.5em}
\label{tab:scenario_2_results}
\end{table}

\begin{figure}[t]
	\centering
	\subfloat{\includegraphics{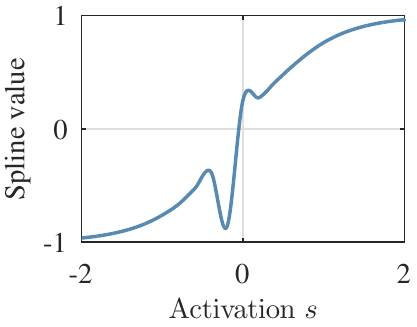}%
		\label{fig:scenario_2_chemical_spline_1}}
	\subfloat{\includegraphics{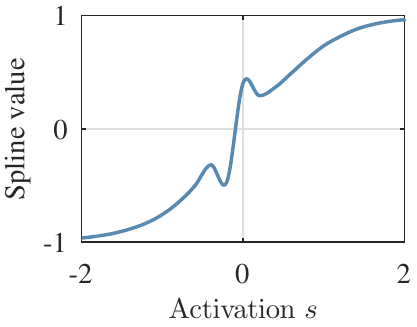}%
		\label{fig:scenario_2_chemical_spline_2}}
	\subfloat{\includegraphics{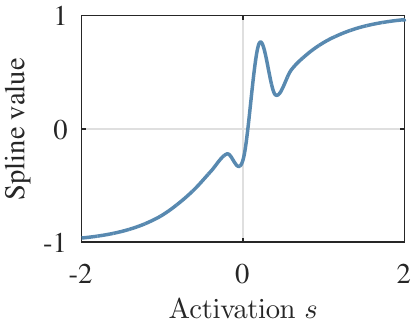}%
		\label{fig:scenario_2_chemical_spline_3}}
	\vfil
	\subfloat{\includegraphics{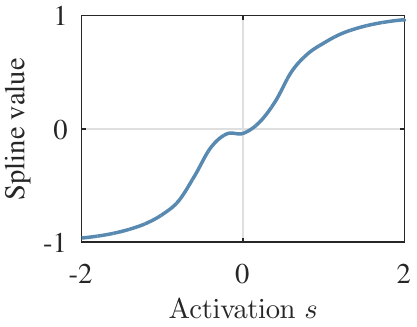}%
			\label{fig:scenario_2_chemical_spline_4}}
	\subfloat{\includegraphics{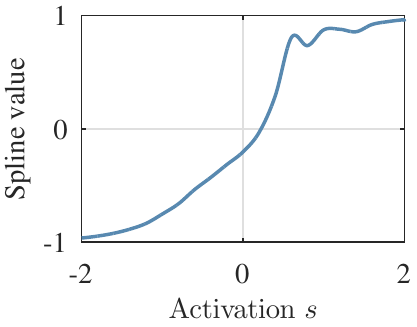}%
		\label{fig:scenario_2_calhousing_spline_1}}
	\subfloat{\includegraphics{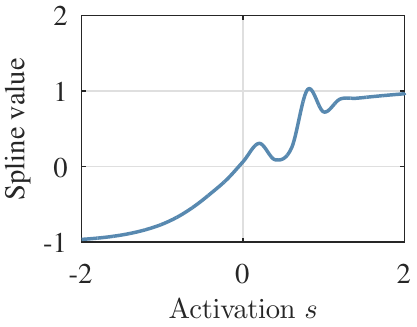}%
			\label{fig:scenario_2_calhousing_spline_2}}
	\caption{Non-trivial representative SAFs after training for scenario $2$.}
	\label{fig:scenario_2_splines}
\end{figure}

Overall, we see that the NNs endowed with the SAF nonlinearities are able to surpass by a large margin a standard NN, and the results from the previous scenario. The only minor drawback evidenced in Table \ref{tab:scenario_2_results} is that the SAF network has some overfitting occurring in the `chemical' dataset (around $2$ points of NRMSE), showing that there is still some room for improvement in terms of spline optimal regularization.

Also in this case, we plot some representatives SAFs after training (taken among those which are not trivially identical to the $\tanh$ nonlinearity) in Fig. \ref{fig:scenario_2_splines}. As before, in general SAFs tend to provide an amplification (with a possible change of sign) of their activation around some region of operation. It is interesting to observe that, also in this case, the optimal shape need not be symmetric (e.g. Fig. \ref{fig:scenario_2_chemical_spline_1}), and might even be far from centered around $0$ (e.g. Fig. \ref{fig:scenario_2_chemical_spline_3}). Resulting nonlinearities can also present some additional non-trivial behaviors, such as a small region of insensibility around $0$ (e.g. Fig. \ref{fig:scenario_2_chemical_spline_4}), or a region of pre-saturation before the actual $\tanh$ saturation (e.g. Fig.s \ref{fig:scenario_2_calhousing_spline_1}-\ref{fig:scenario_2_calhousing_spline_2}).

\section{Conclusion}
\label{sec:conclusion}

In this paper, we have presented a principled way to adapt the activation functions of a neural network from training data, locally and independently for each neuron. Particularly, each nonlinearity is implemented with cubic spline interpolation, whose control points are adapted in the optimization phase. Overfitting is controlled by a novel $\ell_2$ regularization criterion avoiding unwanted oscillations. Albeit efficient, this criterion does constrain the shapes of the resulting functions by a certain degree. In this sense, the design of more advanced regularization terms is a promising line of research. Additionally, we plan on exploring the application of SAFs to deeper networks, where it is expected that the statistics of the neurons' activations can change significantly layer-wise \cite{glorot2010understanding}.

\bibliographystyle{splncs03}
\bibliography{IEEEabrv,../refs}

\end{document}